\documentclass[preprint,12pt]{elsarticle}
\usepackage{hyperref}
\usepackage{url}
\usepackage{natbib}
\usepackage{graphicx}
\usepackage{amsmath,amssymb,latexsym}
\usepackage{multirow}
\usepackage[applemac]{inputenc}

\usepackage{xspace}
\usepackage{tabularx}
\usepackage{multicol}
\usepackage{url}
\usepackage{wrapfig}
\usepackage[normalem]{ulem}
\usepackage{xcolor}

\newcommand{\bb}{\mbox{\boldmath $b$}}
\newcommand{\bc}{\mbox{\boldmath $c$}}

\newcommand{\boldf}{\mbox{\boldmath $f$}}

\newcommand{\bh}{\mbox{\boldmath $h$}}
\newcommand{\bi}{\mbox{\boldmath $i$}}

\newcommand{\bo}{\mbox{\boldmath $o$}}

\newcommand{\br}{\mbox{\boldmath $r$}}

\newcommand{\bx}{\mbox{\boldmath $x$}}

\newcommand{\bz}{\mbox{\boldmath $z$}}

\newcommand{\bU}{\mbox{\boldmath $U$}}
\newcommand{\bV}{\mbox{\boldmath $V$}}
\newcommand{\bW}{\mbox{\boldmath $W$}}
\newcommand{\bX}{\mbox{\boldmath $X$}}

\newcommand{\be}{\begin{eqnarray}}
\newcommand{\ee}{\end{eqnarray}}
\newcommand{\bee}{\begin{eqnarray*}}
\newcommand{\eee}{\end{eqnarray*}}

\newcommand{\matrixb}{\left[ \begin{array}}
\newcommand{\matrixe}{\end{array} \right]}

\journal{Neurocomputing}

\begin{document}

\begin{frontmatter}
\title{Persistent Hidden States and Nonlinear Transformation for Long Short-Term Memory} 

\author{Heeyoul Choi}

\address{Handong Global University \\
558 Handong-ro, Pohang, South Korea, 37554 \\
\texttt{heeyoul@gmail.com} \\
}

\begin{abstract}
Recurrent neural networks (RNNs) have been drawing much attention with great success in many applications like speech recognition and neural machine translation. Long short-term memory (LSTM) is one of the most popular RNN units in deep learning applications. LSTM transforms the input and the previous hidden states to the next states with the affine transformation, multiplication operations and a nonlinear activation function, which makes a good data representation for a given task. The affine transformation includes rotation and reflection, which change the semantic or syntactic information of dimensions in the hidden states. However, considering that a model interprets the output sequence of LSTM over the whole input sequence, the dimensions of the states need to keep the same type of semantic or syntactic information regardless of the location in the sequence. In this paper, we propose a simple variant of the LSTM unit, persistent recurrent unit (PRU), where each dimension of hidden states keeps persistent information across time, so that the space keeps the same meaning over the whole sequence. In addition, to improve the nonlinear transformation power, we add a feedforward layer in the PRU structure. In the experiment, we evaluate our proposed methods with three different tasks, and the results confirm that our methods have better performance than the conventional LSTM. 
\end{abstract}


\end{frontmatter}

\section{Introduction}
In recent deep learning applications like speech recognition, neural machine translation and image caption generation, recurrent neural networks (RNNs) have been successfully applied with great success \cite{Graves2013, Bahdanau2015, Xu2015show}. 
However, it has been known that it is hard to train RNNs, since the gradient information might explode or vanish if it is not carefully controlled in training procedures \cite{Pascanu2012}. To train RNNs efficiently, several approaches have been proposed including long short-term memory (LSTM) or gated recurrent unit (GRU) \cite{Hochreiter1997, Cho2014}. Instead of applying gates for recurrent units, there are other approaches including initialized RNNs (iRNNs) or unitary RNNs (uRNNs) \cite{Le2015, Arjovsky2015}, which focus on the scale of the recurrent connections. 

As a gate based RNN, LSTM is one of the most popular RNN units in deep learning applications and it has several variants for specific tasks \cite{Graves2013, Danihelka2016, Kalchbrenner2015}. LSTM has several advantages based on the gates. First, it reduces the vanishing or exploding gradient problem with the gates, which in turn is very effective for memorizing past information for as long as it is necessary. Second, it transforms the input and the previous hidden states to the next states in a nonlinear fashion, which makes a good data representation for a given task. Note that the affine transformation includes rotation and reflection, which could change the semantic or syntactic information of dimensions in the hidden states across the sequence (or time). 
 
However, considering that a model interprets the output sequence of LSTM over the whole input sequence, the dimensions of the states need to keep the same type of semantic or syntactic information regardless of the location in the sequence. Note that the output of LSTM can be considered as distributed representation where each dimension has its own role \cite{hchoi2017csl, hchoi2018neuro}. For example, in neural machine translation, the attention mechanism is applied to the outputs from the encoder RNN to obtain the context, which is a weighted sum of the outputs. When we calculate the average of vectors with different semantic or syntactic information, the meaning of the average varies. In other words, the dimensions of the states are supposed to have the same meaning. 

In this paper, we focus on the two roles of LSTM: memorization of past information and nonlinear transformation. For memorization, we show that the LSTM structure is not optimized to directly memorize the past information, even though LSTM works well with many applications. As stated above, in LSTM, when the memory cells are updated, the previous memory information is transformed by the affine transformation and then added to the update rule after a nonlinear activation function. We propose removing the affine transformation for the previous hidden states, so that the past information can be delivered to the current state directly. In fact, the cells in LSTM do not need the affine transformation to store the past information. This proposed variant is referred to as {\em persistent recurrent unit (PRU)}. For the nonlinear transformation, since we removed one nonlinear step for the previous hidden states, we add one nonlinear step as a feedforward layer in the PRU structure, which leads to {\em PRU+}. 

In the experiment, we evaluate the proposed recurrent units (PRU and PRU+) for three different tasks: (1)  adding and copying, (2) language modeling, and (3) neural machine translation. The experiments reveal that the proposed units improve memorization and nonlinear transformation compared to the conventional LSTM. 

\section{Background}
In this section, we briefly review conventional RNNs, LSTM and their variants, and discuss how the past information flows into the current state, which is the main issue in this paper. 

\subsection{Recurrent Neural Networks}

In the conventional RNNs, given input data $\bX=[\bx_1, \bx_2, \cdots, \bx_T]$ of length $T$, where each $\bx_t \in R^d$ is a $d$-dimensional vector, the inference for the hidden state at time $t$, $\bh_t$, can be obtained as follows. 
\be
\label{eq:rnn}
\bh_t &=& tanh(\bW \bx_t + \bU \bh_{t-1} + \bb),
\ee
where $\bW$ and $\bU$ indicate the weight matrixes, and $\bb$ is the bias.  Here, $\bh_{t-1}$ makes contribution to $\bh_t$ through the affine transformation which changes the meaning of dimensions across the sequence. Note that, the vanishing or exploding gradient problem happens due to the recurrent connections $\bU$. When the determinant of $\bU$ is greater than 1, the gradient information can explode, and when it is smaller than 1, it can vanish with a long temporal sequence. That is, the problem happens unless the determinant of $\bU$ is 1. To avoid such problems, uRNNs make sure that the determinant stays 1 during the training process \cite{Arjovsky2015}, and  LSTM and GRU deal with the problem with gates, which control the amount of past information flowing into the current state. 

\subsection{Long Short-Term Memory}
\begin{figure}[!h]
\centerline{\hbox{ \includegraphics[width=3.0in]{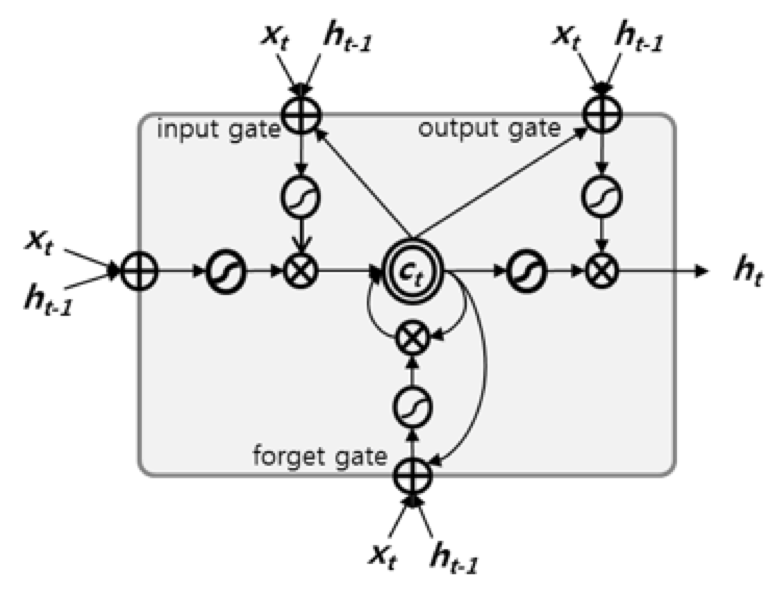}}}
\caption{The structure of a LSTM memory block. Adapted from \cite{Moon2015}.}
\label{fig:lstm}
\end{figure}

In LSTM \cite{Hochreiter1997}, the vanishing or exploding problem can be reduced with three gates: input, forget, and output gates. In order to effectively learn sequential dynamics from the training data, LSTM designs a memory block inside a hidden node, which has a memory cell storing information about the past. The three gates control the flow of past information based on multiplication operations. Fig. \ref{fig:lstm} shows the structure of a single LSTM memory block.

The inference for the hidden space has more complexity than conventional RNNs, and Eq. (\ref{eq:lstm}) shows the exact equations for the inference.  
\be
\label{eq:lstm}
\bc_t &=& \boldf_t\odot \bc_{t-1} + \bi_t\odot tanh(\bU \bh_{t-1} +\bW \bx_t + \bb), \nonumber \\
\bh_t &=& \bo_t\odot tanh(\bc_t),
\ee
where $\bc_t$ and $\bh_t$ are memory cell and hidden states at time $t$, and $\odot$ an element-wise multiplication. Note that $\bh_{t-1}$ is a part of the update rule for the memory cell, $\bc_t$ after the affine transformation, which mixes the meaning of the dimensions, so that LSTM still has the same problem as found in the conventional RNNs. 

Here, we have the three gates, denoted by $i_t$, $f_t$, and $o_t$, respectively, and defined as follows:
\be
\bi_t &=& \sigma( \bW_i x_t+\bU_i \bh_{t-1}+\bV_i \bc_{t-1} + \bb_i), \nonumber \\
\boldf_t &=& \sigma( \bW_f x_t+\bU_f \bh_{t-1}+\bV_f \bc_{t-1}+\bb_f),\nonumber \\
\bo_t &=& \sigma( \bW_o x_t+\bU_o \bh_{t-1}+\bV_o \bc_{t}+\bb_o),
\label{eq:gates}
\ee
where $\sigma$ is the sigmoid function. LSTM learns $\bW$, $\bU$, $\bV$ and $\bb$ for the gates from the training data so that it can determine when to receive input signals, output the hidden node activations, and reset the cell states to refresh the memory. 

Note that for implementation convenience, we do not use the peephole connections $\bV_i \bc_{t-1}, \bV_f \bc_{t-1}$, and $\bV_o \bc_{t}$ which was originally proposed in \citep{Gers2002jmlr}. The peephole connections let the gates look at the cell state to determine the gate values, but the connections are not critical part in LSTM as proved in \cite{Greff2015}, where many different variants to LSTM are presented and compared. 

In addition, the initial states and cells, $\bh_0$ and $\bc_0$, can be trainable parameters, instead of setting to zero, although usually zero initialization is used. In our experiments especially for neural machine translation, there was no significant performance gain with trainable $\bh_0$ and $\bc_0$.

\subsection{Other Approaches} 
While LSTM has been wildly used in the deep learning community, gated recurrent units (GRUs) was proposed especially for language related tasks like machine translation \cite{Cho2014}. GRUs can be considered as a simplified version of LSTM \cite{Greff2015}, and is defined as follows. 
\be
\br_t &=& \sigma( \bW_r x_t+\bU_r \bh_{t-1} + \bb_r), \nonumber \\
\bz_t &=& \sigma( \bW_z x_t+\bU_z \bh_{t-1} + \bb_z), \nonumber \\
\bh_t &=& (1-\bz_t) \odot \bh_{t-1} + \bz_t \odot tanh(\bW x_t + \bU (\br_t \cdot \bh_{t-1})).
\ee
GRUs are different from LSTM in a few ways: (1) there is no peephole, (2) there are only 2 gates, (3) there is no cell state, and (4) there is no output activation. However, the previous information $\bh_{t-1}$ is added to $\bh_{t}$ through the nonlinear transformation. 

Contrary to gated recurrent units like LSTM or GRUs, unitary RNNs take a direct approach to avoid the vanishing (and/or exploding) gradient problems in recurrent neural networks \cite{Arjovsky2015}, and make the determinant of $\bU$ to be 1, which makes $\bU$ be a unitary matrix. However, $\bU$ still rotates and reflects $\bh_{t-1}$ before interacting with $\bW \bx_t$, which changes the meaning of the dimensions in the hidden states. 

To our best knowledge, the previous recurrent units have a nonlinear transformation process of the past information $\bh_{t-1}$ which is integrated to the current information $\bh_{t}$. In the next section, we argue that this process is not necessary for recurrent units, especially in the LSTM architecture, and we propose a new recurrent unit. 

\section{Proposed Structure: Persistant Recurrent Units}
As stated above, conventional RNNs, LSTM, GRU, and unitary RNN have the same issue that the past information $\bh_{t-1}$ is transformed in a nonlinear way to be added to the current information $\bh_{t}$, and such transformation is not necessary to memorize long term dependency. Moreover it makes hard to understand the dimensions of hidden states. 

Even in the unitary RNNs, the unitary matrix, $\bU$ might rotate and/or reflect. To prevent $\bU$ changing the meaning of dimensions in the hidden states, we can add another constraint that the matrix should not rotate nor reflect $\bh_{t-1}$, and the recurrent connection matrix in conventional RNNs becomes an identity matrix as follows.
\be
\bh_t = tanh(\bW \bx_t + \bh_{t-1} + \bb).
\ee 

Moreover, considering the gates in LSTM, the forget gate $\boldf_t$ controls how much past information can be accepted into the cell. That is, $\bc_{t-1}$ is added directly to $\bc_{t}$, and there is no need for the past information $\bh_{t-1}$ in the $tanh$ function. The past information can be delivered directly to $\bc_t$ though $f_t \odot \bc_{t-1}$. Thus, the final equation becomes as follows, coined with {\em persistent recurrent unit} (PRU).  
\be
\label{eq:pru}
\bc_t &=& f_t \odot \bc_{t-1} + i_t \odot tanh(\bW \bx_t + \bb),\nonumber\\
\bh_t &=& o_t \odot tanh(\bc_t),
\ee
where the gate equations are the same as the original LSTM equations as in Eq. (\ref{eq:gates}) without the peephole connections. By removing $\bU \bh_{t-1}$, the units can memorize the past information more easily. In $\bc_t$ of PRU, the past information is only coming from $\bc_{t-1}$ directly, which is enough to keep past information. In LSTM, however, $\bc_t$ is a nonlinear function of $\bh_{t-1}$ as well as $\bc_{t-1}$, where the nonlinear function should keep the past information, and should learn the evolving meaning of the dimension in the hidden states.

In PRU, we removed one step for the nonlinear transformation of the past information to memorize the past information more easily. Now, we focus on the effect of nonlinear transformation in PRU while keeping memorization efficient. Although PRU presents nonlinear transformation, we want to check if the transformation is powerful enough for complex tasks. As a simple approach, we insert one feed-forward layer into the recurrent unit, following $o_t \odot tanh(\bc_t)$. Therefore, a more powerful recurrent unit is proposed, which leads to PRU+, where the cells and the states are updated as follows.  
\be
\bc_t &=& f_t \odot \bc_{t-1} + i_t \odot tanh(\bW \bx_t + \bb),\nonumber\\
\hat{\bh}_t &=& o_t \odot tanh(\bc_t),\nonumber\\
\bh_t &=& tanh(\bW_o \hat{\bh}_t + \bb_o),
\label{eq:pru+}
\ee
where $\bW_o$ is initialized as an identity matrix. In PRU+, the hidden states keep the same dimensions at different time frames and has a more nonlinear transformation. Note that the nonlinear transformation in Eq. (\ref{eq:pru+}) seems less powerful than original LSTM, because the past cell state goes to the current cell state directly not taking the additional nonlinear transformation step. The addition nonlinear transformation in PRU+ is effective only for the output of the unit and the gates for the next step. 

In terms of computational cost, PRU has only one less number of maxtrix-vector multiplication and addition (i.e., $+ \bU \bh_{t-1}$). In GPU environments, however, we could not see any advantage in terms of run-time, because multiple matrix-vector multiplications and additions can be implemented efficiently in parallel. 

To sum up, there are two major roles in RNNs. The first one is memorizing the past information, and the second one is nonlinear transformation as done by other deep neural networks. PRU has advantages in memorizing since the past information does not go though nonlinear function entangled with input $\bx_t$. While directly memorizing the past information, another nonlinear transformation is introduced to the output $\bh_t$ in PRU+. 

\section{Experiments}

Compared to LSTM, we evaluate the performance of our proposed methods (PRU and PRU+) for three different tasks: (1) adding and copying as described in \cite{Arjovsky2015}, (2) language modeling on Penn TreeBank data \cite{Marcus1993penntree} preprocessed by \cite{Mikolov2010} and on the Wikipedia text data in \cite{Merity2016},  and (3) neural machine translation between English and Finish (En-Fi and Fi-En), and between English and Deutch (En-De). We compare the RNN units without adaptive initial states for $\bh_0$ and $\bc_0$, that are set to zeros. Actually, in our experiments the adaptively learnable initial states did not make a significant difference over the zero initial states.

For optimization, we used Adam \cite{Kingma2014adam}, for adaptive learning rate adjustment with gradient clipping (threshold at -1.0 and 1.0). We initialized all the recurrent connection matrices to an identity matrix.

\subsection{Adding and Copying Tasks}
To evaluate how good RNN units are at remembering past information or long-term dependency, we test the proposed methods and LSTM on adding and copying tasks. To generate data sets for the two tasks, we followed \cite{Arjovsky2015, Hochreiter1997}. The models have only one RNN layer.

The data set for the adding task is generated as follows. Each input has two sequences of length $T$. The first one consists of numbers sampled uniformly between 0 and 1. The second sequence indicates two locations where the two source numbers are. The first source entry is located randomly in the first half of the sequence, while the second entry is in the second half. The output is the sum of the two entries of the first sequence. 

The goal is to memorize the locations for the two numbers and add them. The cost function is the mean squared error (MSE) between the true calculation and the estimated sum by the model. If the algorithm predicts 1 regardless of the input sequence, the expected mean squared error is 0.167, which is the baseline for this task. The network structures are the same with 128 hidden units for the all RNN units. 

\begin{figure}[!h]
\centerline{\hbox{ \includegraphics[width=2.7in]{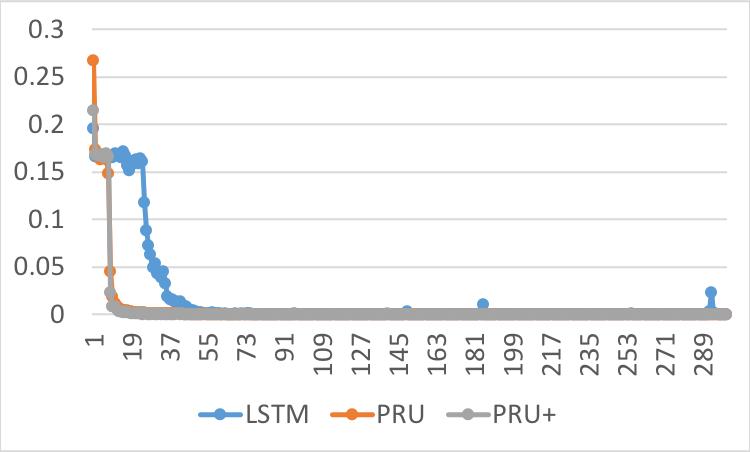}}\hbox{ \includegraphics[width=2.7in]{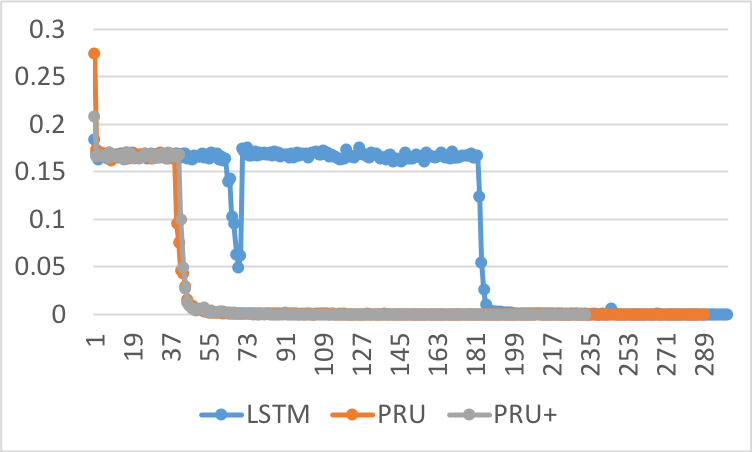}}}
\caption{Mean Squared Error (MSE) for the adding task with three different recurrent units for different sequence lengths. 
(Left) length=100, (Right) length=500. The horizontal axis indicates the number of epochs for training.}
\label{fig:adding}
\end{figure}

As shown in Fig. \ref{fig:adding}, PRU and PRU+ converge faster than LSTM. This is because PRU and PRU+ can keep the past information without the rotation and reflecting operations. That is, PRU can learn quickly how to select and memorize relevant information. For this task, there is no significant difference between PRU and PRU+, because this task focuses on memorizing, not on nonlinear transformation. Note that as the length of the sequence increases, the gap between the number of iterations for convergence increases. 

The copying task considers 10 categories, $C_0, C_1, \cdots, C_9$, for 8 data categories and 2 control categories. Each input takes a sequence of categories with length of $T + 20$. The first 10 entries are randomly sampled categories from 8 data categories, $C_0$ to $C_7$, which is to be memorized. The next $T - 1$ entries are sequence of $C_8$. The next single entry is $C_9$, which indicates the starting time to reproduce the memorized 10 categories in the output. The last 10 entries are $C_8$. The output sequence should be $T + 10$ $C_8$ and the first 10 categories in the input sequence. That is, the first 10 entries should be memorized and reproduced after $T$ steps. 

The cost function for the copying task is the average of the cross entropy of predictions at each time step of the sequence. A simple memoryless strategy is to predict $C_8$ for the first $T+10$ steps and random categories from $C_0$ to $C_7$ for the last 10 steps. 
The cost of the simple memoryless strategy is $\frac{10 log(8)} {T+20}$ as in \cite{Arjovsky2015}. 

\begin{figure}[!h]
\centerline{\hbox{ \includegraphics[width=2.7in]{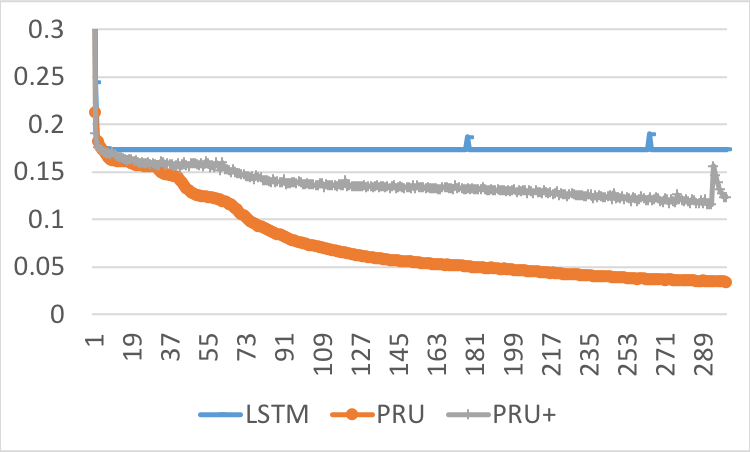}} \hbox{ \includegraphics[width=2.7in]{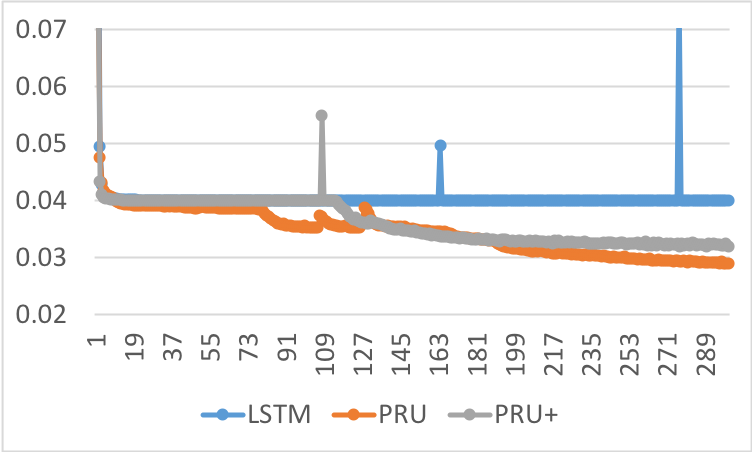}}}
\caption{Mean Squared Error (MSE) for the copying problem with three recurrent units for different sequence lengths. (Left) length=100, (Right) length=500. The horizontal axis indicates the number of epochs for training.}
\label{fig:copying}
\end{figure}

Fig. \ref{fig:copying} shows that PRU and PRU+ converge faster than LSTM, which seems to not converge at least within 300 iterations, which is the same as in \cite{Arjovsky2015}. Note that PRU is faster than PRU+, which means the nonlinear transformation seems not necessary, though the task looks more complicated to learn than the adding task. 

From the adding and copying tasks, we could see that PRU and PUR+ learn to memorize input data more quickly than LSTM. 

\subsection{Language Model}

In addition to the synthesized tasks above (adding and copying), we evaluate our RNN units with more realistic tasks like language modeling, which probably need more complicated nonlinear transformations. Since the first neural network based language model was proposed \cite{Bengio2003}, RNNs have been widely adapted for language models \cite{Mikolov2011}, which is a crucial part of speech recognition or neural machine translation \cite{Moon2015, gulcehre2015using}. Also, language modeling is one of the most popular tasks to evaluate RNN performance \cite{Karpathy2015}. 

We trained character-level language models using Penn TreeBank data and word level language models using  the Wikipedia text data (WikiText-103-raw) in \cite{Merity2016} with different RNN units. Each input sequence is a sequence of characters (or words) in one sentence. The number of hidden nodes is set to 1000 for both language models. The cost function is the average of the cross entropy values over the sequence of the prediction. 

Table. \ref{table:lm} summarizes the experiment results in terms of negative log-likelihood (NLL) for the Penn TreeBank data. Since nonlinear transformation might be crucial to a language model, we included LSTM+ which is implemented in the same way as PRU+, except that it still has the term $\bU \bh_{t-1}$ for the cell state update. From the table, we can see LSTM+ is not helping which might imply that LSTM already has enough power for nonlinear transformation. However, PRU has better performance compared to LSTM, and PRU+ is almost the same as (if not slightly better than) PRU. It may mean that PRU is successful in memorizing information and transforming it, and PRU+ can add slightly more nonlinear transformation effect without losing memorization efficiency. In other words, PRU is so simple that it can effectively memorize past information, so it improves the performance by only memorizing well. Then, PRU+ improves the performance by adding nonlinear transformation power on top of the advantage of PRU. 

Also, note that for test data, PRU and PRU+ generalize better than LSTM and LSTM+, and PRU is the best, probably because the term $\bU \bh_{t-1}$ makes the model overfitting and is not necessary to learn temporal dynamics in the RNN units. 

\begin{table}[!h]
    \centering
\begin{tabular}{l||c|c|c|}
Models & Validation NLL (V) & Test NLL (T) & Generalization (V-T)\\
\hline
LSTM       & 102.37 & 101.39 & 0.98\\
LSTM+ & 102.91 & 101.96 & 0.95\\
PRU  	& 101.53 & 100.22 & {\bf 1.31}\\
PRU+ 	& {\bf 101.06} & {\bf 99.79} & 1.27 \\
\hline 
\end{tabular}
\caption{Negative log-likelihood of language models for Penn TreeBank data}
\label{table:lm}
\end{table}


To see how the proposed methods work with large-scale data, we added language modeling task with Wikipedia Text data where the training data size is 516MB. Table. \ref{table:lm_wiki} summarizes NLL with different lengths of sentences. From the table, we can see that when the length is short (less than or equal to 50), the RNN units have the same (or very similar) performance. However, when the length is large, the proposed methods outperform the conventional LSTM. It indicates that the proposed units can keep past information better than LSTM. Also, the nonlinear transformation is not crucial for word-level language model. 

\begin{table}[!h]
    \centering
\begin{tabular}{l||c|c||c|c|c|}
& \multicolumn{2}{c||}{Sentence Length $\leq$ 50} & \multicolumn{2}{c|}{Length $\leq$ 250} \\
\hline
Models & Valid NLL & Test NLL & Valid NLL & Test NLL \\
\hline
LSTM       & 51.11 & 44.03 & 431.00 & 398.32 \\
PRU  	& 51.15 & {\bf 43.97} & 417.31 & {\bf 387.68} \\
PRU+ 	& 51.65 & 44.24 & 422.21 & 392.25 \\
\hline 
\end{tabular}
\caption{Negative log-likelihood of language models for the Wikitext-103 dataset}
\label{table:lm_wiki}
\end{table}

\subsection{Neural Machine Translation}

As another real-world task, neural machine translation is designed based on RNNs \cite{Sutskever2014,Bahdanau2015}. In this section, we evaluate the three RNN units with neural machine translation, where the same RNN unit is used for both the encoder and decoder of a model.  

\subsubsection{Models}
As a neural machine translation model, we use the attention-based neural translation model from \cite{Bahdanau2015} as a baseline, except for replacing GRU with LSTM. Also, in the experiments, the vocabulary size is 10K for En-Fi (and Fi-En), and 30K for En-De. For both the source and target languages, we used the same vocabulary size, and the same dimension of word embedding, which is 300. The number of the hidden nodes for both the encoder and decoder is 500, and the dimension of the hidden nodes for the alignment model is 1000. Based on the model configuration above, we use the same kind of RNN units among LSTM, PRU, and PRU+ for both the encoder and decoder. 

All the models are trained using Adam~\cite{Kingma2014adam} until the BLEU score on the validation set stops improving. To measure the validation score during training, we use greedy search (1 for the beam width) instead of beam search in order to minimize the computational overhead. As in \cite{Bahdanau2015}, we trained our models with sentences that contain up to 50 words. 

\subsubsection{Tasks and Corpora}

Our evaluations are based on three translation tasks; (1) En-Fi, (2) Fi-En, and (3) En-De. We use the parallel corpora available from WMT'15\footnote{\url{http://www.statmt.org/wmt15/}} for training, which results in around 2M sentence pairs for En-Fi and 4.5M pairs for En-De. We do not use any preprocessing routine other than simple tokenization.

Instead of space-separated words, we use 10K (or 30K) subwords extracted by byte pair encoding (BPE), as suggested in \cite{Sennrich2016bpe}. When computing the translation quality using BLEU, we {\it un-BPE} the resulting translations, but leave them tokenized.

\subsubsection{Decoding and Evaluation}

Once a model is trained, we use a simple forward beam search with the width set to 1 or 12 to find a translation that approximately maximizes $\log p(Y|X)$, where $X$ and $Y$ refer to source and target sentences, respectively. The decoded translation is then un-BPE'd and evaluated against a reference sentence by BLEU (in practice, BLEU is computed over a set of sentences.). As the validation and test sets, we use `newsdev2015' and `newstest2015' for En-Fi, and `newstest2013' and `newstest2015' for En-De. Tables. \ref{table:bleu} and \ref{table:bleu_de} represent the performance with the three RNN units with different data configurations. We can see that PRU is as good as LSTM, though PRU is slightly better than LSTM for En-Fi and En-De. As in the language model experiment, we can see that the computation power of PRU is almost the same as LSTM. This might be a combination of more efficient memorization and less powerful nonlinear transformation. The term $\bU \bh_{t-1}$ for the cell state may not be necessary in neural machine translation. Moreover, PRU+ is significantly better than LSTM because it adds the nonlinear transformation effect on top of better memorization. 

\begin{table}[t]
    \centering
\begin{tabular}{l||c|c|c|c|c|c}
& \multicolumn{2}{c|}{En-Fi} & \multicolumn{2}{c|}{Fi-En} \\
\hline
Beam Size 	&  1 & 12 			& 1 & 12 \\ 
\hline
\hline
\small LSTM		&   7.31 (7.83) &  8.82 (9.22)  & 10.2 (11.38) & 11.93 (13.02) \\
\small PRU		&  7.69 ( 8.13) & 9.18 (9.81)  & 10.06 (11.32) & 11.81 (13.08)\\ 
\small PRU+         &  {\bf 8.14} ({\bf 8.68}) & {\bf 9.64} ({\bf 10.1}) & {\bf 10.77} ({\bf 11.83}) & {\bf 12.26} ({\bf 13.32})\\
\hline
\hline
\end{tabular}
\caption{BLEU scores on the test sets for En-Fi and Fi-En with two different beam widths. The scores on the development sets are in the parentheses. The LSTM model is the vanilla model from \cite{Bahdanau2015} with LSTM and BPE.}
\label{table:bleu}
\end{table}%

\begin{table}[t]
    \centering
\begin{tabular}{l||c|c|c|c}
\hline
Beam Size 	&  1 & 12 			\\ 
\hline
\hline
\small LSTM		&   19.97 (19.26) &  22.63 (21.12)   \\
\small PRU		&  19.94 (19.41) & 22.74 (21.00) \\ 
\small PRU+         &  20.04 (19.32) &  22.98 (21.05) \\
\hline
\hline
\end{tabular}
\caption{BLEU scores on the test sets for En-De with two different beam widths. The scores on the development sets are in the parentheses.}
\label{table:bleu_de}
\end{table}%

\begin{figure}[!h]
\centerline{\hbox{ \includegraphics[width=2.7in]{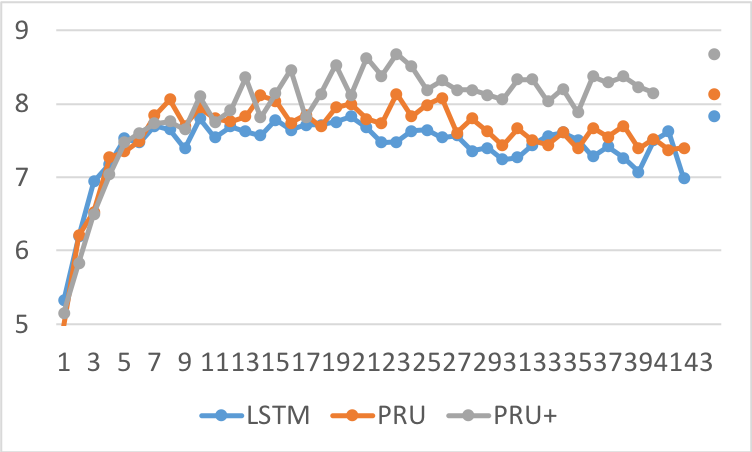}} \hbox{ \includegraphics[width=2.7in]{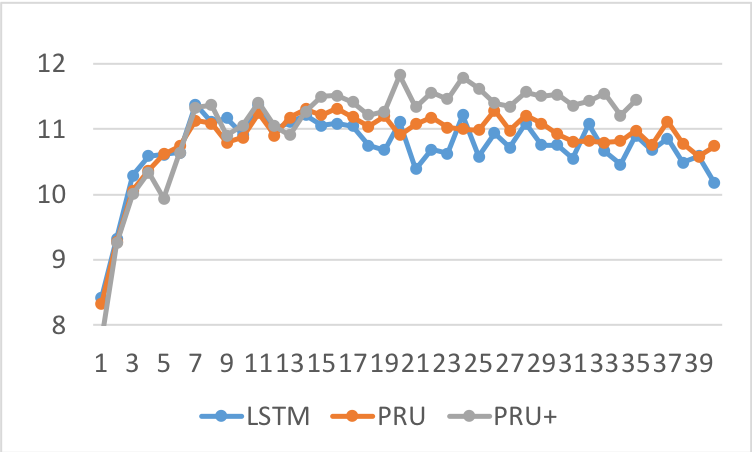}}}
\caption{BLEU scores of validation data during training with the beam width 1. (Left) En-Fi, (Right) Fi-En. The horizontal axis indicates the number of iterations ($\times5000$) for training, and the vertical axis
BLEU scores.}
\label{fig:nmt}
\end{figure}

There is another interesting property in PRU+. In Fig. \ref{fig:nmt}, PRU+ does not have the overfitting problem. Considering the size of data set for En-Fi or Fi-En, usually there is the overfitting issue, and LSTM and PRU overfit. Obviously, PRU+ has a similar model structure and slightly more parameters than PRU, but PRU+ does not overfit, while PRU does. Although we need more studies to understand these results exactly, our current interpretation is that the feedforward steps prevent overfitting because the transformation increases the volume of the output so that the outputs easily saturate by the $tanh$ function. It is known that when the outputs of the nonlinear activation function are saturated, the loss function can probably converge more on a flat region rather than a sharp one \cite{Hochreiter1997flat, Chaudhari2016}, and the flat region provides better generalization for test data. Actually after training the models where $\bW_o$ is initialized as an identity matix, the average of the diagonal elements of $\bW_o$ for the PRU+ units is 3.34 \footnote{The determinant of the $500\times500$ matrix is $\inf$ in python implementation.}, which can push the outputs by the transformation to the saturation areas.

\begin{figure}[!h]
\centerline{\hbox{ \includegraphics[width=5.5in]{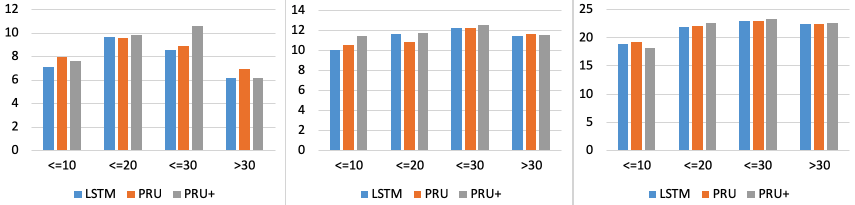}}}
\caption{BLEU scores for the different lengths of sentences for the test data of three translation tasks: (Left) En-Fi, (Middle) Fi-En, and (Right) En-De. The horizontal axes indicate the range of sentence length.}
\label{fig:bleu_length}
\end{figure}

To check how the proposed units learn long-term dependences, we evaluate the BLEU score with different length of sentences. In Fig. \ref{fig:bleu_length}, BLEU scores are presented with different length of sentences for the three NMT tasks. PRU and PRU+ are as good as (or slightly better than) original LSTM to learn long-term dependency. 

\section{Conclusions}
We understand LSTM in two aspects: (1) memorizing the past information, and (2) transforming data nonlinearly. The hidden states are transformed into the next states with the affine transformation before the nonlinear function. The affine transformation mixes the semantic or syntactic meanings. However, the dimensions of the states are supposed to have the same meaning. Also, the nonlinear transformation of the previous hidden states to the next state is not necessary to memorize the past information.  
  
Based on such understanding, we proposed a simple variant of the LSTM unit, PRU. In PRU, the dimension of the hidden states are persistent across time, so that the hidden space keeps the same meaning over the whole sequence. This helps the units memorize the past information more easily. In addition, to empower PRU, we added a feedforward layer in the PRU structure, which provides more powerful nonlinear transformation. In the experiment results, PRU and PRU+ outperform the conventional LSTM, which confirms our understanding of the recurrent units. In future work, the gates can be analyzed and improved in a similar fashion.

\section*{Acknowledgment}
This research was supported by Basic Science Research Program through the National Research Foundation of Korea(NRF) funded by the Ministry of Education (2017R1D1A1B03033341), and by Institute for Information \& communications Technology Promotion(IITP) grant funded by the Korea government(MSIT) (No. 2018-0-00749, Development of virtual network management technology based on artificial intelligence).

\section*{References}
\bibliographystyle{elsarticle-num}

\end{document}